\documentclass[lettersize,journal]{IEEEtran}
\usepackage{amsmath,amsfonts}
\usepackage{algorithmic}
\usepackage{algorithm}
\usepackage{array}
\usepackage[caption=false,font=normalsize,labelfont=sf,textfont=sf]{subfig}
\usepackage{textcomp}
\usepackage{stfloats}
\usepackage{url}
\usepackage{verbatim}
\usepackage{graphicx}
\usepackage{cite}
\usepackage{amssymb}
\usepackage{booktabs}
\usepackage{multirow}
\usepackage{bbm}
\usepackage{color,xcolor}
\usepackage{empheq}
\usepackage{float}
\newtheorem{definition}{Definition}

\begin{document}

\title{Knowledge-Assisted Multi-Graph Dependency Learning for\\ Multivariate Time Series Anomaly Detection in\\ Multi-Stage Industrial Processes}

\author{{Jaeyeong~Lee, Taeseong~Yoon, Wonmo~Koo, and Heeyoung~Kim}
\thanks{This work was supported by the National Research Foundation of Korea (NRF) grant funded by the Korea government (MSIT) (2023R1A2C2005453, RS-2023-00218913). (Corresponding author: Heeyoung Kim.)}
\thanks{Jaeyeong Lee, Taeseong Yoon, Wonmo Koo, and Heeyoung Kim are with the Department of Industrial and Systems Engineering, Korea Advanced Institute of Science and Technology (KAIST), Daejeon, Republic of Korea. (e-mail: dlwodud116@kaist.ac.kr; bigstar0423@kaist.ac.kr; joseph92@kaist.ac.kr; heeyoungkim@kaist.ac.kr).}
\thanks{Jaeyeong Lee and Taeseong Yoon contributed equally to this work.}}

\markboth{IEEE Transactions on Automation Science and Engineering}
{Lee \MakeLowercase{\textit{et al.}}: Knowledge-Assisted Multi-Graph Dependency Learning for Multivariate Time Series Anomaly Detection}

\maketitle

\begin{abstract}
Industrial processes often generate complex, interdependent time-series data from multiple sensors across multiple stages, forming complex dependencies among variables and process stages. Effective monitoring and timely anomaly detection of these time series through multivariate time series anomaly detection (MTAD) is crucial for preventing failures and ensuring the reliability of automated systems. 
Graph neural networks (GNNs) have advanced MTAD by leveraging data-driven graphs to model complex dependencies among variables, effectively capturing relational structures within multivariate time series to enhance anomaly detection performance.
However, existing GNN-based approaches often overlook critical process knowledge, and even when this knowledge is considered, seamlessly incorporating it into existing models remains inherently challenging, leading to suboptimal performance.
To address this limitation, we propose a knowledge-assisted multi-graph framework for modeling sensor dependencies in multi-stage industrial processes for MTAD, which explicitly incorporates process knowledge into graph learning to enhance dependency modeling and improve anomaly detection performance.
Our method constructs three complementary graphs: one purely data-driven and two refined by integrating structural constraints derived from process knowledge.
To effectively leverage these graphs for anomaly detection, we employ a multi-graph attention network, enabling a more accurate and robust representation of complex dependencies. Comprehensive experiments on two real-world, multi-stage industrial datasets demonstrate that incorporating process knowledge substantially enhances anomaly detection performance.
\\

\textit{Note to Practitioners---}This work tackles anomaly detection in multi-stage industrial processes, where many sensors exhibit dependencies both within and across stages. Methods that learn sensor relations purely from data often overlook critical dependencies or infer spurious ones. To address this limitation, the proposed approach incorporates two forms of commonly available process knowledge---sensor group knowledge (sensors associated with each sub-process) and process flow knowledge (adjacent sub-process relationships)---derived from process diagrams.
The method builds three complementary graphs (uninformed, sensor group knowledge-informed, and process flow knowledge-informed) and uses a multi-graph attention network combined with a temporal convolutional network for forecasting-based anomaly detection. Experiments on water treatment and water distribution systems show substantial F1 improvements over data-driven baselines, ablation studies confirming the contributions of both knowledge sources.
For deployment, practitioners only need to specify sensor groupings and stage adjacencies from existing diagrams; no manual relationship modeling or specialized domain expertise is required. The model is trained on normal-operation data and detects anomalies via prediction errors. The approach is most effective when sub-process boundaries and flows are well documented and should be updated if the plant configuration changes. Beyond water systems, this method can be applied to chemical processing, power generation, and automated manufacturing facilities with similar multi-stage structures.
\end{abstract}


\section{Introduction}
\label{sec:introduction}
\IEEEPARstart{I}{ndustrial} control systems (ICS) are cyber-physical systems designed to enable a small number of operators to efficiently manage industrial processes. These systems rely on numerous sensors and actuators distributed across multiple stages (or sub-processes) to monitor and control operations \cite{mathur2016swat, sharma2016overview, park2022prediction, ma2022smart,cho2023prediction}. 
Effective monitoring and timely anomaly detection of the multivariate time series generated by ICS are important for preventing failures and ensuring system reliability, which can be achieved through multivariate time series anomaly detection (MTAD).

Recently, methods employing graph neural networks (GNNs) \cite{zhao2020multivariate,deng2021graph,chen2021learning,chen2022deep,han2022learning,dai2022graph,shi2023robust,ding2023mst,jo2024edge,liu2024topogdn,jeong2026pgrf} have shown remarkable performance in MTAD, primarily due to their ability to capture complex inter-sensor dependencies through graphs, where nodes represent sensors and edges represent their relationships. Since the structure of the graph is often unknown in many real-world MTAD scenarios \cite{deng2021graph}, GNN-based MTAD methods typically learn the graph structure from the data.

\begin{figure*}[t]
    \centering
    \subfloat[]{\includegraphics[width=3.5in]{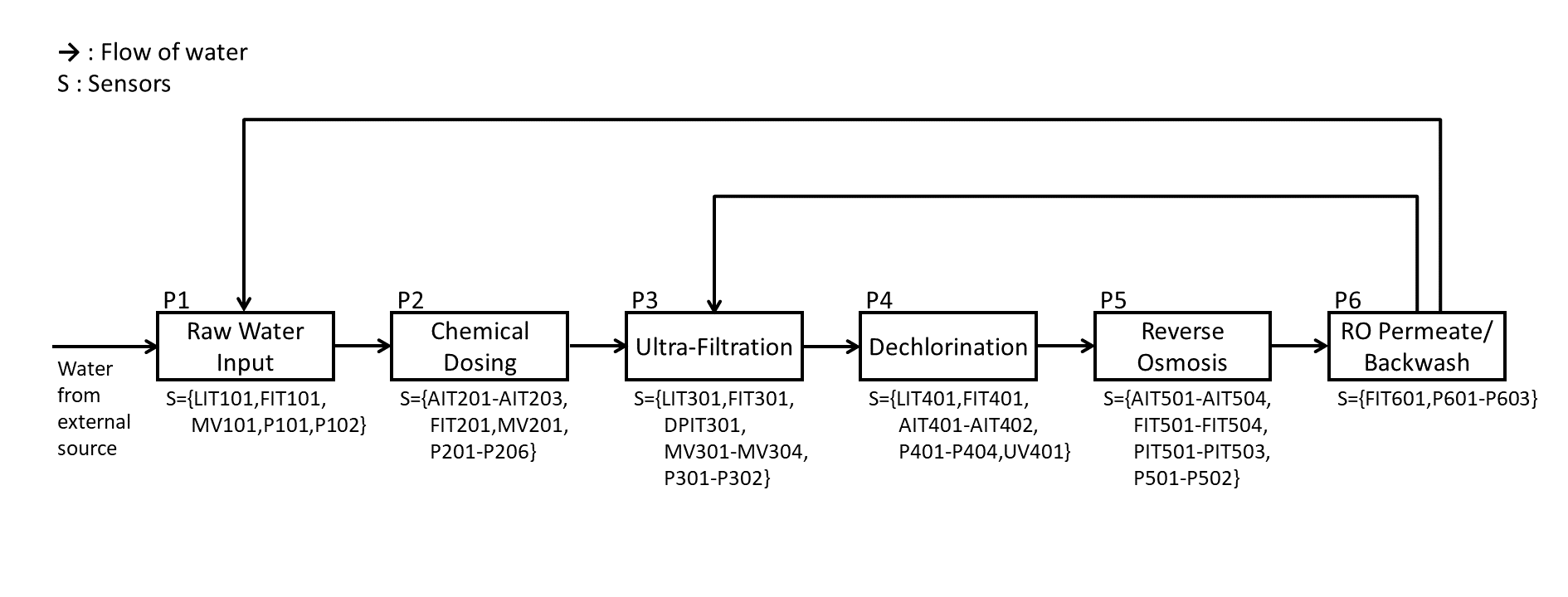}\label{fig: figure 1a}}
    \subfloat[]{\includegraphics[width=3.5in]{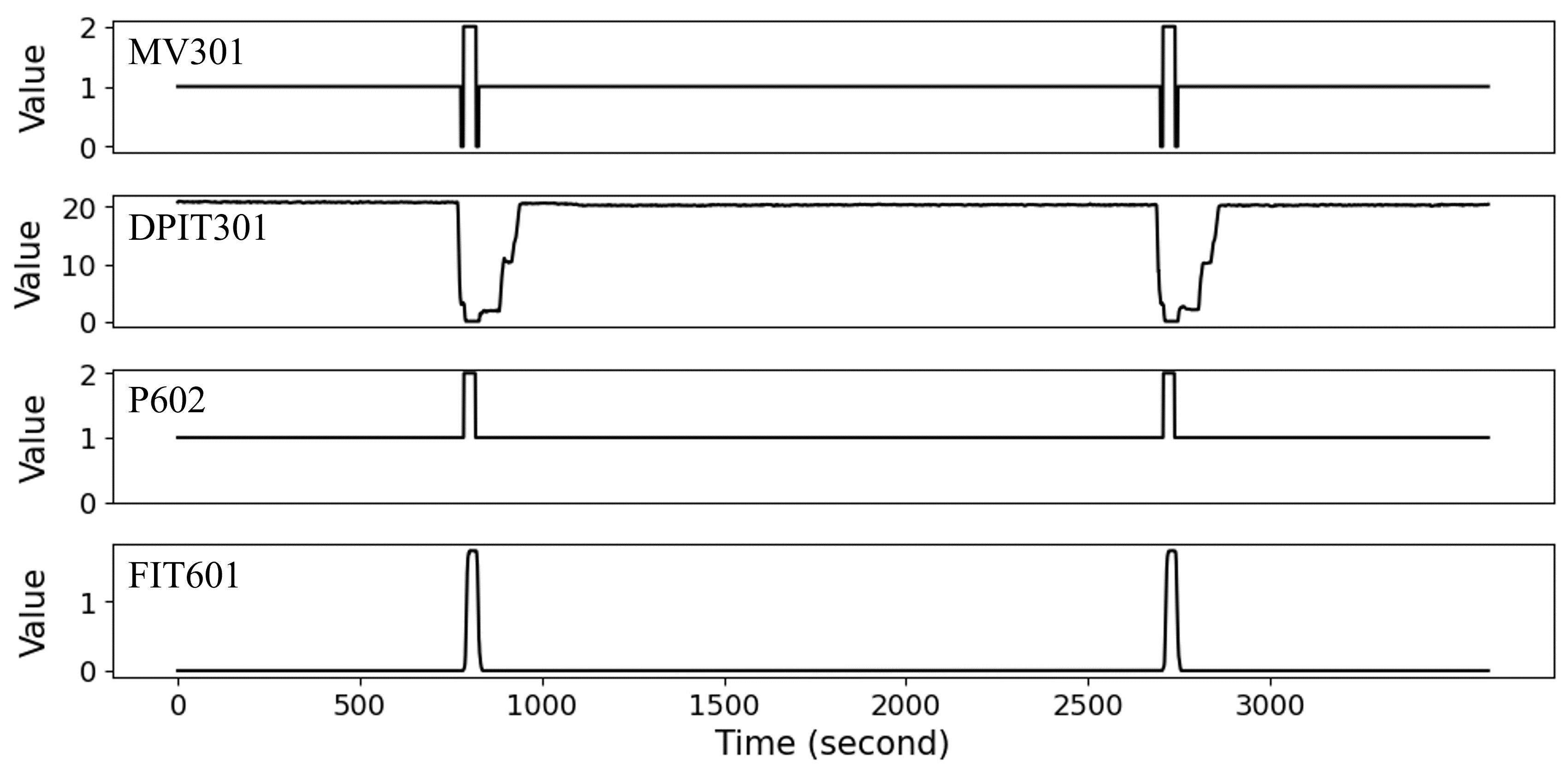}\label{fig: figure 1b}}
    \caption{(a) The process flow diagram of the Secure Water Treatment (SWaT) process and (b) a one-hour-long multivariate time series from four sensors in P3 and P6.}
    \label{fig: figure 1}
\end{figure*}

However, accurately learning the graph structure can be challenging for processes with multiple stages or sub-processes, which exhibit complex dependencies both among variables and across process stages. In particular, strong correlations can arise not only among variables within the same stage but also between different stages, especially when they are closely linked \cite{ma2026order}. 
To illustrate this challenge, Fig.~\ref{fig: figure 1}(a) presents the process flow diagram of a water treatment process comprising six sub-processes (P1--6), each monitored by multiple sensors, 
while Fig.~\ref{fig: figure 1}(b) highlights the strong correlations among the measurements from two sensors in P3 (MV301 and DPIT301) and two sensors in P6 (P602 and FIT601). These correlations naturally arise from the underlying process dynamics. Specifically, in P3, MV301 regulates the backwashing valve for the ultra-filtration (UF) unit, while DPIT301 measures the resulting pressure difference. Similarly, in P6, P602 indicates the pump's state, directly influencing the water flow measured by FIT601. Additionally, P3 and P6 are intrinsically linked through the backwashing process, as treated water from P6 is pumped to P3 to clean the UF unit.
\footnote{Although P3 and P6 are sequentially connected in process flow, their sensor responses may appear nearly simultaneous in Fig.~\ref{fig: figure 1}(b) because the physical response delay is short relative to sampling resolution. Thus, the figure illustrates coupled responses from sequentially connected sub-processes, rather than simultaneous operation of unrelated sub-processes.}

\begin{figure*}[t]
 \centerline{\includegraphics[width=0.7\linewidth]{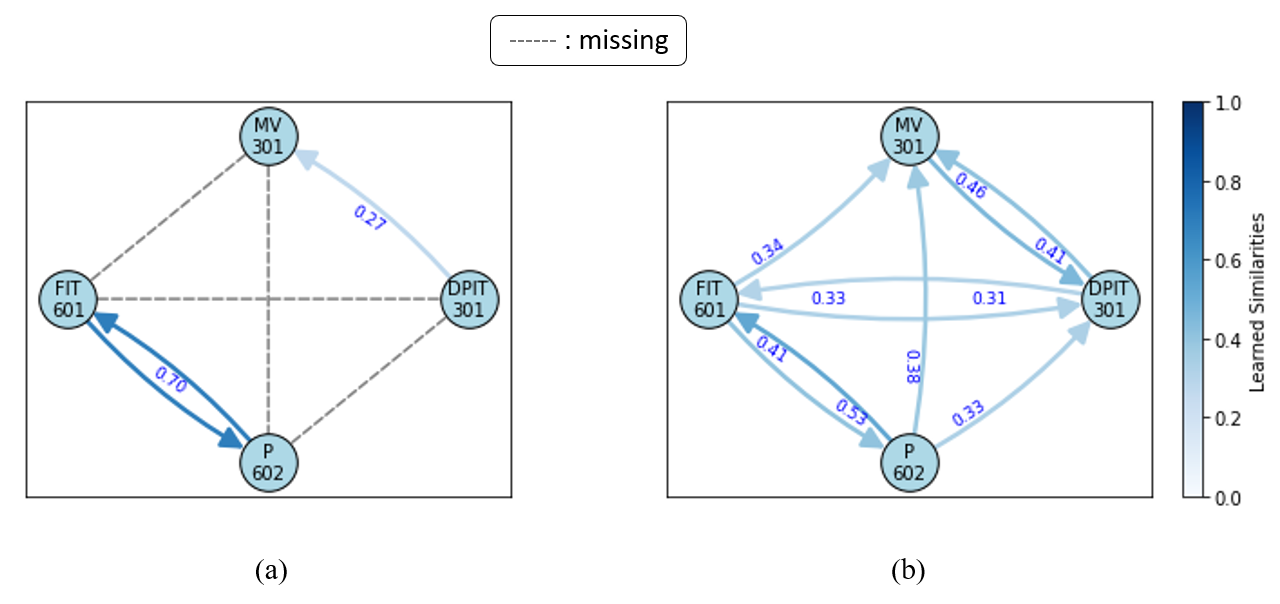}}
 \caption{Relationships among four sensors captured through graph structure learning in (a) the GDN \cite{deng2021graph} and (b) the proposed model. Edges are weighted by learned similarities, with dashed lines representing missing relationships.}
 \label{fig: toy example}
\end{figure*}

Thus, accurately modeling these complex interdependencies is crucial for effective MTAD. 
However, existing GNN-based MTAD methods, which infer inter-sensor relationships solely from data, often struggle to accurately capture these complex dependencies. As a result, they may learn spurious connections or fail to identify key relationships in multi-stage processes. 
This limitation is illustrated in Fig.~\ref{fig: toy example}(a), which displays the relationships among the four sensors in Fig.~\ref{fig: figure 1}(b) (MV301, DPIT 301, P602, FIT601) learned by the graph deviation network (GDN) \cite{deng2021graph}, a representative GNN-based MTAD method. Solid lines indicate detected relationships, while the dashed lines represent missing connections. The numerical values along the edges denote similarity scores, with darker shades indicating stronger inferred connections. 
Ideally, the model should capture all relationships among four sensors, given their high correlations, as shown in Fig.~\ref{fig: figure 1}(b). However, GDN misrepresents sensor interactions, leading to an incomplete depiction of dependencies and ultimately suboptimal anomaly detection.
In contrast, our proposed method, shown in Fig.~\ref{fig: toy example}(b), effectively captures the true sensor dependencies, leading to more reliable anomaly detection.

To address these challenges, sensor group knowledge---which specifies the sub-process associated with each sensor---has been incorporated into a GNN-based MTAD framework \cite{han2022learning}.
While this approach improved dependency modeling by leveraging sensor group knowledge, it did not account for \textit{process flow knowledge}, which captures relationships between closely related sub-processes. Without explicitly modeling relationships between interconnected sub-processes, the learned graph structure may still overlook critical dependencies, ultimately limiting the effectiveness of anomaly detection.
Thus, a more comprehensive approach that integrates both sensor group knowledge and process flow knowledge is essential for accurately modeling multi-stage industrial processes and enhancing MTAD performance.

In this paper, we propose a knowledge-assisted multi-graph framework for modeling sensor dependencies in multi-stage industrial processes for forecasting-based MTAD. 
The framework extends GNN-based MTAD approaches to structurally incorporate sensor group knowledge and process flow knowledge into graph construction, and exploits the resulting complementary relational representations through multi-graph attention for forecasting.
This structured integration helps the model to more effectively capture inter-sensor dependencies, improving over conventional data-driven approaches that often overlook the underlying process dynamics. 
By coupling this knowledge-guided graph structure construction with temporal modeling via a temporal convolutional network (TCN) \cite{bai2018empirical}, the proposed method delivers more robust forecasting and improved anomaly detection performance on real-world industrial datasets.

Our method consists of three stages: (1) graph structure learning, (2) graph attention-based forecasting, and (3) anomaly scoring. 
Our primary contribution lies in the first stage---graph structure learning---where we introduce a principled method to embed process knowledge into the graph.
This is nontrivial due to architectural constraints and the difficulty of embedding structured relationships into graph representations.

To address this, we introduce a simple yet effective structural design that integrates process knowledge seamlessly into the GNN framework.
Specifically, we represent sensor similarities using two learnable embeddings per sensor, capturing bidirectional relationships between sensor pairs.
Based on these similarities, we construct three graphs: (1) a purely data-driven graph, (2) a graph that selectively masks similarities using sensor group knowledge, and (3) a graph that masks similarities based on process flow knowledge.
Instead of merely treating prior knowledge as auxiliary node attributes, our framework translates it directly into the graph construction mechanism.
This introduces structured inductive bias into graph learning itself, guiding the robust discovery of inter-sensor dependencies.

Building upon these graphs, we forecast future sensor readings and conduct anomaly detection based on forecast errors. 
A key challenge here is to integrate the learned inter-sensor dependencies from graphs with intra-sensor temporal dynamics to enable accurate forecasting and robust anomaly detection.
Since each graph encodes distinct inter-sensor relationships, we employ three separate graph attention networks (GATs) \cite{velickovic2017graph}, one for each graph constructed in the previous stage, each learning an attention mechanism tailored to the corresponding graph and enabling the model to capture complementary relational information for forecasting. 
These GATs allow for flexible and adaptive relationship modeling by dynamically assigning varying levels of importance to neighboring sensors.
In parallel, to model intra-sensor dependencies over time, we use a TCN, leveraging its ability to effectively capture temporal patterns with relatively low computational cost compared to typical recurrent models.

The outputs of the GATs and TCN are combined to form rich spatiotemporal representations, which are used to forecast future sensor readings. 
Anomaly scores are then derived from the resulting prediction errors.
We validate our method on two real-world multi-stage industrial datasets and show that incorporating process knowledge significantly improves anomaly detection performance.

The main contributions of this paper are threefold. 
(1) a knowledge-assisted multi-graph dependency learning framework for MTAD in multistage industrial processes;
(2) a principled mechanism for modeling both intra-sub-process and inter-sub-process dependencies
using sensor group knowledge and process flow knowledge; and
(3) comprehensive empirical validation, including ablation and qualitative analyses, demonstrating the effectiveness and interpretability of the proposed knowledge-guided graph design.

The remainder of the paper is organized as follows. Section \ref{sec:review} reviews related work. Section \ref{sec:background} provides an overview of the GAT and TCN. Section \ref{sec:methodology} describes the proposed method. In Section \ref{sec:experiments}, we validate the proposed method using real-world sensor datasets. Finally, we conclude this paper in Section \ref{sec:conclusion}.

\section{Related Work}
\label{sec:review}
\subsection{Unsupervised Multivariate Time Series Anomaly Detection}

Following the standard taxonomy of anomaly detection, the present work focuses on point anomaly detection rather than contextual or collective anomaly detection \cite{chandola2009anomaly,choi2021deep,kim2023dpvae, kim2023contextual}.

For MTAD, various deep neural network (DNN)-based methods have been proposed \cite{choi2021deep, zamanzadeh2022deep, kim2023deep, li2023deep}, owing to their effectiveness in handling complex, high-dimensional time series data. 
Due to the unlabeled nature of the data and the rarity of anomalies, MTAD has mainly been treated as an unsupervised learning problem \cite{chen2022deep}. Unsupervised DNN-based MTAD methods are typically categorized into two classes: forecasting-based approaches \cite{hundman2018detecting, wu2020developing, chen2021learning, liu2023spacecraft,kim2023time} and reconstruction-based approaches \cite{li2019mad, su2019robust,audibert2020usad, thill2021temporal, tuli2022tranad,zhang2023stad}.

Forecasting-based MTAD methods typically predict the value at each time stamp by using previously observed values and use forecasting errors to detect anomalies. These methods operate under the assumption that forecasting errors of anomalous data points are higher than those of normal data points. To capture complex intra-sensor dependencies, recurrent neural networks (RNNs) have been used \cite{hundman2018detecting,wu2020developing}. Furthermore, TCNs and Transformers \cite{vaswani2017attention} have also been employed due to their effectiveness in capturing long-range intra-sensor dependencies \cite{chen2021learning,liu2023spacecraft,kim2023time}.

Reconstruction-based MTAD methods typically reconstruct observed time series, identifying anomalies when reconstruction errors are significant. Similar to forecasting-based MTAD methods, they employ various DNNs specially designed to capture complex intra-sensor dependencies, such as RNNs \cite{park2018multimodal,li2019mad,su2019robust,zhang2023stad}, TCNs \cite{thill2021temporal}, and Transformers \cite{wang2022variational,tuli2022tranad}. Beyond such intra-sensor modeling, \cite{dai2024sarad} applied a Transformer to capture inter-sensor spatial associations. \cite{wu2025catch} performed reconstruction in the frequency domain by patchifying the spectrum into bands, and adaptively discovered channel correlations between channel-independent and channel-dependent strategies. More recently, diffusion-based generative models such as \cite{zhong2025multi} have also been explored for reconstruction-based anomaly detection.

Hybrid MTAD methods \cite{zhao2020multivariate,chen2022deep,han2022learning,guan2022gtad,ding2023mst} combine forecasting and reconstruction models. These methods commonly use both forecasting and reconstruction errors to detect anomalies. More recently, contrastive learning has been explored as an alternative paradigm, learning discriminative representations through synthetic anomaly injection or causality-aware augmentation \cite{darban2025carla, pmlr-v267-kim25aa}.

\subsection{Graph Neural Network-Based MTAD}
In recent years, several GNN-based MTAD methods \cite{zhao2020multivariate,deng2021graph,chen2021learning,chen2022deep,han2022learning,dai2022graph,shi2023robust,ding2023mst,jo2024edge,liu2024topogdn,jeong2026pgrf} have been introduced, achieving remarkable performance by effectively capturing complex inter-sensor dependencies as well as intra-sensor dependencies. In an initial effort, \cite{zhao2020multivariate} used a GAT \cite{velickovic2017graph}, assuming that sensor relationships could be represented by a complete graph. However, in many real-world cases, the assumption of a complete graph is unrealistic, and the underlying (sparse) graph is typically unknown \cite{deng2021graph}. To address this challenge, \cite{deng2021graph} proposed a GNN-based MTAD method capable of learning the latent graph from the data. They introduced a randomly initialized learnable embedding for each sensor to represent its characteristics, deriving the graph from the similarities across the sensor embeddings. This graph structure learning technique has been employed in various GNN-based MTAD methods in a similar manner \cite{chen2022deep,han2022learning,ding2023mst,jo2024edge,liu2024topogdn}. In addition to sensor embedding-based techniques, various graph structure learning techniques have been employed in GNN-based MTAD methods. For example, \cite{chen2021learning} learned a graph by edge sampling, employing the Gumbel-softmax sampling trick \cite{jang2016categorical} to make the sampling procedure differentiable. Furthermore, \cite{dai2022graph} learned a directed acyclic graph (DAG) through gradient-based optimization with a differentiable constraint \cite{zheng2018dags}. Beyond learning the graph in an end-to-end manner, \cite{febrinanto2025entropy} constructed the sensor graph from transfer entropy to explicitly capture causal relationships among sensors. However, most GNN-based MTAD methods may exhibit limitations in performance when applied to multi-stage industrial processes, as they overlook the distinctive characteristics of such processes, which can be encoded as sensor group knowledge and process flow knowledge.   
Recently, \cite{han2022learning} extended the method by \cite{deng2021graph} to better suit multi-stage industrial processes. They incorporated sensor group knowledge into the process of learning sensor embeddings. 
However, they did not consider process flow knowledge.
In contrast, our work incorporates process flow knowledge in addition to sensor-group knowledge within a unified multi-graph construction framework, enabling explicit modeling of both intra-sub-process and inter-sub-process dependencies.

\section{Background}
\label{sec:background}

\subsection{Temporal Convolutional Network}
\label{subsec:TCN}
TCNs \cite{bai2018empirical} are a class of one-dimensional convolutional neural networks designed to capture long-range temporal dependencies in multivariate time series data. 
TCNs are built from stacked \textit{dilated causal convolutional} layers.
The \textit{causality} constraint ensures that the output at time $t$ depends only on input at time $t$ or earlier,
while \textit{dilation} introduces fixed gaps between filter taps, allowing the receptive field to grow without increasing filter size.

\begin{figure}[htbp]
    \centerline{\includegraphics[width=\columnwidth]{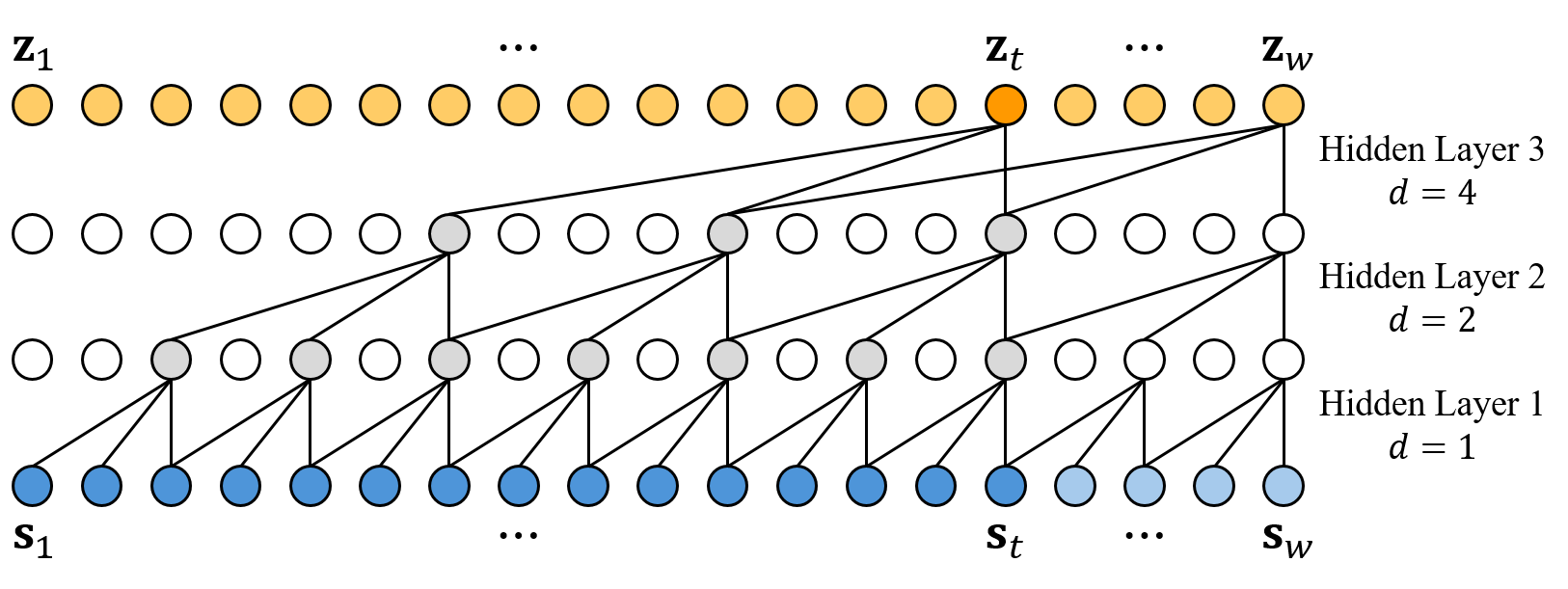}
    }
    \caption{Illustration of a TCN with three dilated causal convolutional layers.} 
    \label{TCN_overview}
\end{figure}

Formally, for a one-dimensional input sequence $\textbf{s} \in \mathbb{R}^{w}$ and a convolutional filter $f:\{0, \dots, J-1\} \to \mathbb{R}$, where $J$ is the filter size, the output sequence $\textbf{z} \in \mathbb{R}^{w}$ is computed as:
\begin{equation*}
    \label{eq: dilated causal convolution}
    \textbf{z}_{t} = \sum_{j=0}^{J-1}f(j) \cdot \textbf{s}_{t-d\cdot j},
\end{equation*}
where $\textbf{z}_t \in \mathbb{R}$ and $\textbf{s}_t \in \mathbb{R}$ are the value of $\textbf{z}$ and $\textbf{s}$ at time $t$, and $d$ is the dilation factor.
By stacking multiple such layers with exponentially increasing dilation rate (e.g., $d=2^{i-1}$ for the $i$-th layer), TCNs effectively expand their receptive field and capture long-range dependencies.
Fig.~\ref{TCN_overview} illustrates a TCN with three dilated causal convolutional layers, using $d=1,2,4$ and $J=3$.

\subsection{Graph Attention Network}
\label{subsec:GAT}
GATs \cite{velickovic2017graph} are a class of GNNs that utilize self-attention to assign varying importance to neighboring nodes during feature aggregation, improving upon uniform averaging in graph convolutional networks (GCNs) \cite{kipf2016semi} and effectively capturing complex dependencies between different nodes.

Given a graph $\mathcal{G}$ with $N$ nodes and input features $\textbf{z} \in \mathbb{R}^{N \times w}$, GAT computes an updated feature $\textbf{a}_{i} \in \mathbb{R}^{d}$ for node $i$ as follows.
First, each node feature $\mathbf{z}_{i} \in \mathbb{R}^w$ is linearly transformed using a shared weight matrix $\textbf{W} \in \mathbb{R}^{d\times w}$, and attention coefficients $\pi_{ij} \in \mathbb{R}$ are computed via a function $f: \mathbb{R}^{d} \times \mathbb{R}^{d} \to \mathbb{R}$:
\begin{equation*}
    \pi_{ij} = f(\mathbf{W}\mathbf{z}_i,\mathbf{W}\mathbf{z}_j), \quad j \in 
    \begin{cases}
        \mathcal{N}(i)\cup\{i\}, & \text{if self-loop at } i,\\
        \mathcal{N}(i), & \text{otherwise},
    \end{cases}
\end{equation*}
where $\mathcal{N}(i)$ is the set of neighbors of node $i$ in $\mathcal{G}$.
Then, if there is a self-loop at node $i$, the output node feature $\textbf{a}_i$ is calculated as follows:
\begin{equation*}
    \mathbf{a}_i = \sigma(\alpha_{ii}\mathbf{W}\mathbf{z}_i+\sum\limits_{j \in \mathcal{N}(i)} \alpha_{ij}\mathbf{W}\mathbf{z}_j),
\end{equation*}
\begin{equation*}
    \alpha_{ij} = \frac{\exp (\pi_{ij})}{\sum_{r \in \mathcal{N}(i)\cup{\{i\}}}\text{exp}(\pi_{ir})},
\end{equation*}
where $\sigma(\cdot)$ is a nonlinear activation function.

\section{Methodology}
\label{sec:methodology}
\begin{figure*}[h] 
    \centerline{\includegraphics[width=0.7\textwidth]{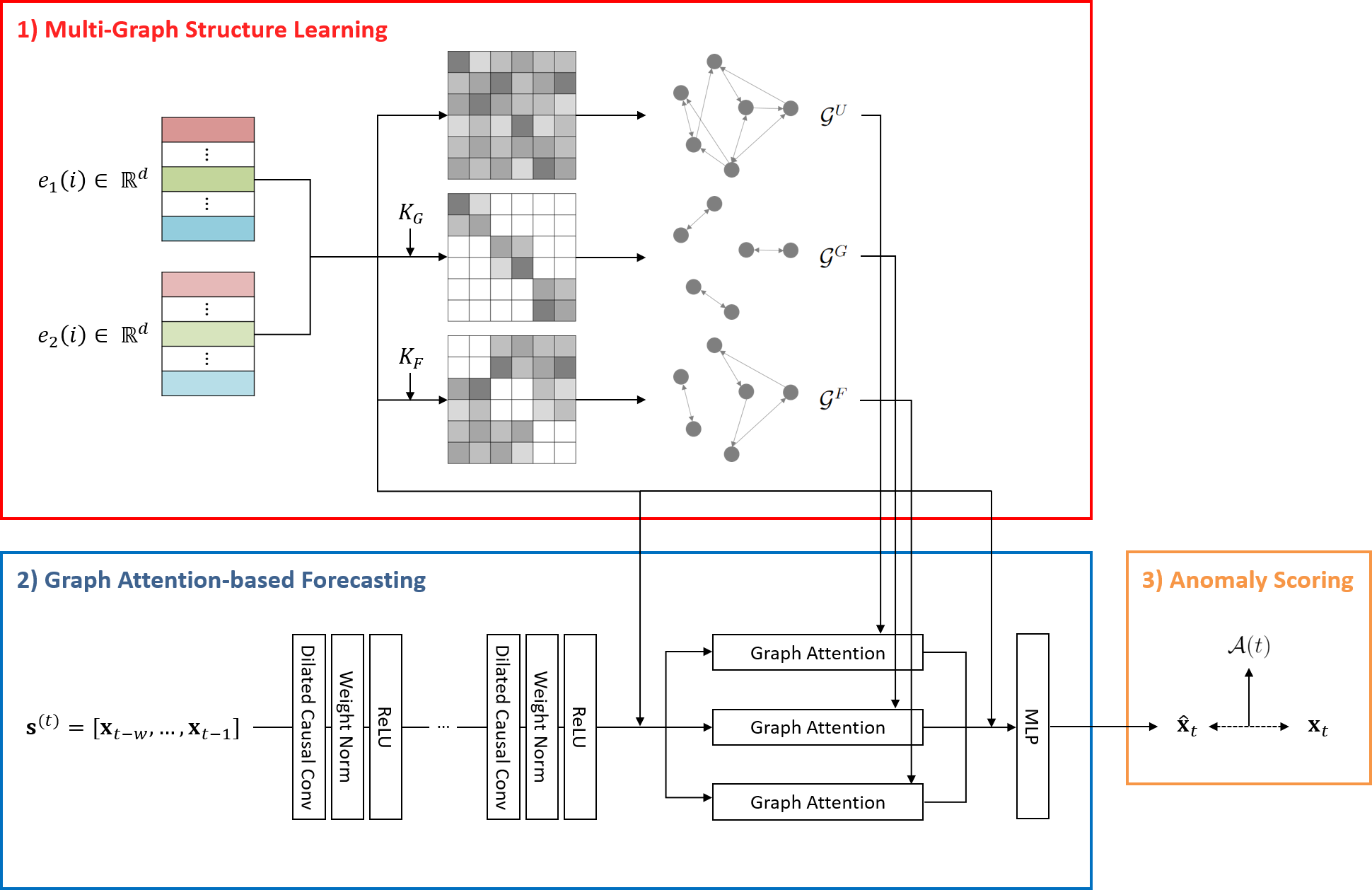}}
    \caption{Overview of the proposed method} 
    \label{overview}
\end{figure*}

In this section, we present a proposed method in detail. Our approach comprises three main components: (1) graph structure learning, (2) graph attention-based forecasting, and (3) anomaly scoring. An overview is provided in Fig.~\ref{overview}. 
Our key contribution lies in enhancing graph structure learning by integrating \textit{process knowledge} to improve dependency modeling.

The remainder of this section is structured as follows. 
First, we define the problem setting of MTAD and formally define process knowledge in Section \ref{problem statement}. 
Next, we present a multi-graph structure learning framework, focusing on our core contributions: the innovative strategy for incorporating process knowledge (Section \ref{subsec:multi-graph structure learning}). Then, we describe the graph attention-based forecasting framework, which effectively leverages the three constructed graphs (Section \ref{subsec:graph attention based forecasting}). Finally, we introduce the anomaly scoring scheme, designed to detect anomalies accurately using the proposed framework (Section \ref{sec:anomaly scoring}). 
The codes for the proposed method are available at \url{https://anonymous.4open.science/r/Knowledge-Assisted-Multi-Graph-Structure-Learning-60AE}.

\subsection{Problem Statement}
\label{problem statement}
We consider multivariate time series data generated from a multi-stage industrial process consisting of $N_p$ sub-processes.  
We denote the collection of sensor index sets for the entire process as $\{P_s\}_{s=1}^{N_p}$, where $P_s$ is the set of sensor indices associated with the $s$-th sub-process. The total number of sensors is given by $N= \sum\limits_{s=1}^{N_p} |P_s|$. These sub-processes form a process flow graph, and we denote their adjacency as a symmetric binary relation $S \subseteq \{P_s\}_{s=1}^{N_p} \times \{P_s\}_{s=1}^{N_p}$, where $(P_{s_1}, P_{s_2}) \in S$ indicates that the $s_1$-th and $s_2$-th sub-processes are directly connected. 
To formally express the sensor-to-sub-process relationship, we define the mapping
\begin{equation}
\label{eq:phi_mapping}
\phi: \{1,\ldots N\} \rightarrow \{1, \ldots N_{p}\}, 
\end{equation}
where, $i \in \{1,\ldots,N\}$ denotes a sensor index, and $\phi(i)$ gives the index of the sub-process to which sensor $i$ belongs.

Each sensor observation $\mathbf{x}_t \in \mathbb{R}^N$ at time $t$ is composed of $N_p$ sub-vectors $\mathbf{x}_t = \left[\mathbf{x}_t^{(1)}; \mathbf{x}_t^{(2)}; \dots; \mathbf{x}_t^{(N_p)}\right]$, where $\mathbf{x}_t^{(s)} \in \mathbb{R}^{|P_s|}$ denotes the sensor values of the $s$-th sub-process at time $t$ associated with the sensor indices in $P_s$. 
Our training dataset consists of $T_\text{tr}$ sensor observations collected under normal operating conditions, without any anomalies. In contrast, the test dataset contains $T_\text{te}$ sensor observations, which may include both normal and anomalous data.

Our goal is to detect anomalies in the test dataset. The proposed method produces $T_\text{te}$ binary labels, each indicating whether the sensor value at a given test timestamp is anomalous (e.g., value 1) or normal (e.g., value 0). 
We focus on point anomalies, where anomalies are defined at the individual timestamp level without explicitly modeling contextual or subsequence-level anomaly patterns.
We define two types of prior knowledge commonly found in multi-stage processes:

\begin{definition}[Sensor group knowledge]
\label{def1}
\textbf{Sensor group knowledge} $K_G = \{P_s\}_{s=1}^{N_p}$ is a collection of sensor index sets, where $P_s$ represents the sensors involved in the $s$-th sub-process.
\end{definition}

\begin{definition}[Process flow knowledge]
\textbf{Process flow knowledge} $K_F = \{F_s\}_{s=1}^{N_p}$ is a collection of sensor index sets, where $F_s=\mathop{\cup}\limits_{r: (P_r, P_s) \in S} P_r$ represents the sensors involved in sub-processes adjacent to the $s$-th sub-process.
\end{definition}

\subsection{Multi-Graph Structure Learning}
\label{subsec:multi-graph structure learning}
By incorporating sensor group knowledge and process flow knowledge, we learn three graphs: 1) uninformed graph $\mathcal{G}^U$, 2) sensor group knowledge-informed graph $\mathcal{G}^G$, and 3) process flow knowledge-informed graph $\mathcal{G}^F$.

\subsubsection{Uninformed Graph}
The uninformed graph $\mathcal{G}^U$ is learned without using any prior knowledge about sensor relationships. We adopt the graph structure learning technique of \cite{deng2021graph}, which is a foundational similarity-based method and a strong baseline. 
To model directional relationships between sensors, we introduce two learnable embeddings for each sensor, representing the source and target roles, respectively. 
The source and target embeddings for each sensor are randomly initialized and jointly learned during training by minimizing the forecasting loss. The embedding dimension $d$ is treated as a hyperparameter.

Specifically, for each sensor $i\in\{1,\ldots, N\}$, we assign two learnable embeddings $e_{1}(i), e_{2}(i) \in \mathbb{R}^{d}$, representing its latent characteristics when acting as a source and target, respectively. 
For a directed relation from sensor $j$ to sensor $i$, sensor $j$ is regarded as the source node, i.e., the candidate influencing sensor, and sensor $i$ as the target node, i.e., the sensor whose dependency is modeled. This design enables the model to capture asymmetric relationships between sensors.
These embeddings $e_{1}(i)$ and $e_{2}(i)$ are initialized randomly and are learned automatically through backpropagation under the forecasting objective.
Thus, they are not externally specified or separately estimated, but are optimized jointly with the rest of the network in an end-to-end manner.
The similarity between the $j$-th sensor (source) and the $i$-th sensor (target) is computed as follows:
\begin{equation*}
    C_{ji}=\text{ReLU}\bigg(\frac{e_1(j)^\top e_2(i)}{||e_1(j)||_2\cdot ||e_2(i)||_2}\bigg),
\end{equation*}
where $||\cdot||_2$ denotes the $L_2$-norm, and ReLU refers to a rectified linear unit \cite{nair2010rectified}. Then, we obtain the adjacency matrix $A^U = (A^U_{ji}) \in \mathbb{R}^{N \times N}$, which represents the uninformed graph $\mathcal{G}^U$ based on the similarities $C_{ji}$, as follows:
\begin{equation}
    \label{eq: A^{U}}
    A^U_{ji} = \mathbbm{1}\{j \in \text{Top}_k(\{C_{ri}| r \in \{1,2,\dots,N\}\text{\textbackslash}\{i\}\})\},
\end{equation}
where $\text{Top}_k$ is an operator that selects the indices of the $k$-highest values among its input and $\mathbbm{1}\{\cdot\}$ is the indicator function, returning $1$ if the condition is true and $0$ otherwise.
The parameter $k$ controls graph sparsity by determining the number of incoming edges per node. 
The value of $k$ is treated as a hyperparameter and is selected on the validation set together with $\alpha$ and $d$ through a joint grid search.

\subsubsection{Sensor Group Knowledge-Informed Graph}
\label{subsubsection:sensor group knowledge graph}
To model intra-sub-process relationships, we use sensor group knowledge $K_G = \{P_s\}_{s=1}^{N_p}$. The sensor group knowledge-informed graph $\mathcal{G}^G$ focuses on dependencies among sensors within the same sub-process. 
We define the masked similarity between the $j$-th sensor (source) and the $i$-th sensor (target) as follows: 
\begin{equation*}
    C_{ji}^{G}  = \begin{cases} C_{ji}, &  \text{if}\, j \in P_{\phi(i)},\\
    -\infty, & \text{otherwise,} \end{cases}
\end{equation*}
where $\phi(\cdot)$ is the sensor-to-sub-process mapping defined in \eqref{eq:phi_mapping}. 
This mapping is constructed based on the predefined sensor group knowledge $K_{G} = \{P_{s}\}_{s=1}^{N_{p}}$.
Then, we obtain the adjacency matrix $A^G = (A_{ji}^{G})\in\mathbb{R}^{N \times N}$, which represents the sensor group knowledge-informed graph $\mathcal{G}^G$ based on the masked similarities $C_{ji}^{G}$, as follows:
\begin{equation}
    A^G_{ji} = \mathbbm{1}\{j \in \text{Top}_{k_i^G}(\{C_{ri}^G| r \in \{1,2,\dots,N\}\text{\textbackslash}\{i\}\})\},
    \label{eq:topk_eq}
\end{equation}
where $k_i^{G} = \alpha \cdot |P_{\phi(i)}|$ and $\alpha \in (0,1)$ is a hyperparameter controlling the sparsity of $\mathcal{G}^{G}$. 
This adaptive neighborhood size reflects the variations in the number of sensors across sub-processes, ensuring that each sensor selects a proportionally consistent number of intra-sub-process neighbors.

\subsubsection{Process Flow Knowledge-Informed Graph}
\label{subsubsection:process flow knowledge graph}
We further incorporate process flow knowledge $K_F = \{F_s\}_{s=1}^{N_p}$ to model inter-sub-process relationships. The process flow knowledge-informed graph $\mathcal{G}^F$ focuses on dependencies among sensors across adjacent sub-processes.
Following a similar masking strategy used in constructing the sensor knowledge-informed graph (Section \ref{subsubsection:sensor group knowledge graph}), we compute the masked similarities as follows:
\begin{equation*}
    C_{ji}^{F}  = \begin{cases} C_{ji}, &  \text{if}\, j \in F_{\phi(i)}, \\ -\infty, & \text{otherwise,} \end{cases}
\end{equation*}
where $\phi(\cdot)$ is the sensor-to-sub-process mapping defined in \eqref{eq:phi_mapping}. 
Here, the condition $j \in F_{\phi(i)}$ indicates that the $j$-th sensor belongs to one of the sub-processes that are adjacent to the $i$-th sub-process, as specified by the process flow knowledge $K_{F} = \{F_{s}\}_{s=1}^{N_{p}}$. 
Then, the adjacency matrix $A^F = (A_{ji}^{F}) \in \mathbb{R}^{N \times N}$, which represents the process flow
knowledge-informed graph $\mathcal{G}^{F}$, is obtained by
\begin{equation}
    A^F_{ji} = \mathbbm{1}\{j \in \text{Top}_{k_i^F}(\{C_{ri}^F| r \in \{1,2,\dots,N\}\text{\textbackslash}\{i\}\})\},
    \label{eq:topk_eq2}
\end{equation}
where $k_i^F = \alpha \cdot |F_{\phi(i)}|$, using the same hyperparameter $\alpha$ as in the construction of  $\mathcal{G}^G$.

This multi-graph approach may offer several advantages. First, maintaining three complementary graph structures can provide alternative pathways for dependency modeling when process knowledge is incomplete or noisy, thereby enhancing robustness against knowledge uncertainties. Furthermore, the distinct contribution of each graph to anomaly detection can be examined, offering insights into the relative importance of different dependency types and thus improving interpretability.

\subsection{Graph Attention-based Forecasting}
\label{subsec:graph attention based forecasting}
We perform graph attention-based forecasting by using the three graphs $\mathcal{G}^U$, $\mathcal{G}^G$, and $\mathcal{G}^F$. At each time $t$, we use a sliding window of size $w$ over historical timestamps, $\mathbf{s}_t:= [\mathbf{x}_{t-w},\dots,\mathbf{x}_{t-1}] \in \mathbb{R}^{N \times w}$ as an input to the proposed method. Our goal is to predict the next observation $\mathbf{x}_t$.

\subsubsection{Temporal Encoder}
The input sequence $\mathbf{s}_t$ is first passed through a TCN to capture intra-sensor dependencies. The TCN consists of multiple dilated causal convolutional layers, each followed by weight normalization \cite{salimans2016weight} and ReLU activation, following \cite{bai2018empirical}. 
The encoded feature $\mathbf{z}_t \in \mathbb{R}^{N\times w}$ is computed via the TCN formulation presented in Section \ref{subsec:TCN}:
\begin{equation*}
    \mathbf{z}_t = \text{TCN}(\mathbf{s}_t).
\end{equation*}

\subsubsection{Multi-Graph Attention Network}
\label{subsubsection: multigat}
To capture inter-sensor dependencies, we apply three separate GATs, each operating over one of the learned graphs $\mathcal{G}^U$, $\mathcal{G}^G$, and $\mathcal{G}^{F}$, respectively.
For each graph, we follow the procedure described in Section~\ref{subsec:GAT} to compute the corresponding aggregated feature vectors $\mathbf{a}_{t,i}^{U} \in \mathbb{R}^d$, $\mathbf{a}_{t,i}^{G} \in \mathbb{R}^d$, and $\mathbf{a}_{t,i}^{F} \in \mathbb{R}^d$, respectively. 
As an example, for the uninformed graph $\mathcal{G}^U$, the aggregated feature $\mathbf{a}_{t,i}^{U} \in \mathbb{R}^d$ for the $i$-th sensor is computed as follows: 
\begin{equation*}
    \mathbf{a}_{t,i}^{U} = \text{ReLU}(\beta_{ii}^{U}\mathbf{W}^{U}\mathbf{z}_{t,i}+\sum\limits_{j \in \mathcal{N}^{U}(i)} \beta_{ij}^{U}\mathbf{W}^{U}\mathbf{z}_{t,i}),
\end{equation*}
\begin{equation*}
    {\beta}_{ij}^{U} = \frac{\text{exp}(\pi_{ij}^{U})}{\sum_{r \in \mathcal{N}^{U}(i)\cup{\{i\}}}\text{exp}(\pi_{ir}^{U})},
\end{equation*}
where $\textbf{z}_{t,i}\in\mathbb{R}^w$ is the $i$-th row of $\textbf{z}_t$, denoting the feature of the $i$-th sensor from the temporal encoder, $\mathbf{W}^{U} \in \mathbb{R}^{d \times w}$ is a weight matrix, and $\mathcal{N}^{U}(i) = \{j|A^U_{ji}=1\}$, where $A^{U}_{ji}$ is defined in \eqref{eq: A^{U}}. 
Notably, attention-based aggregation is explicitly dependent on the structure of $\mathcal{G}^{U}$, as both the neighbor set $\mathcal{N}^{U}(i)$ and the attention weights $\beta_{ij}^{U}$ are computed based solely on the edges defined in the adjacency matrix $A^{U}$.

The attention score $\pi^{U}_{ij}$ is adapted from the formulation in \cite{deng2021graph}, with a key modification; we explicitly distinguish the source and target roles of a sensor by using separate embeddings $e_{1}(\cdot)$ and $e_{2}(\cdot)$. Specifically, 
\begin{empheq}[left=\empheqlbrace]{align*}
    \pi_{ij}^{U} &= \text{LeakyReLU}((\mathbf{u}^{U})^\top(r_{\text{target},i}^{U} \oplus r_{\text{source},j}^{U})), \notag\\
    r_{\text{target},i}^{U} &=e_2(i) \oplus \mathbf{W}^{U}\mathbf{z}_{t,i}, \quad
    r_{\text{source},j}^{U} = e_1(j) \oplus \mathbf{W}^{U}\mathbf{z}_{t,j},
\end{empheq} 
where $W^{U}$ and $u^{U}$ are randomly initialized trainable parameters of the attention module and jointly optimized with the rest of the network during end-to-end training.
In addition, $\oplus$ denotes the concatenation operator, and LeakyReLU refers to the leaky rectified linear unit \cite{xu2015empirical}.  
Here, $r_{\text{target},i}^{U}$ is a contextualized representation of the $i$-th sensor as a target node, while $r_{\text{source}, j}^{U}$ corresponds to the representation of the $j$-th sensor as a source node. Each representation combines structural embeddings ($e_{2}(i)$ or $e_{1}(j)$) with temporal features ($\mathbf{z}_{t,i}$), enabling the model to capture direction-aware dependencies between sensors.
This asymmetric formulation enhances the model's capacity to represent complex and structured inter-sensor interactions.

Aggregated features $\mathbf{a}_{t,i}^{G} \in \mathbb{R}^d$ and $\mathbf{a}_{t,i}^{F} \in \mathbb{R}^d$ are computed analogously using $\mathcal{G}^{G}$ and $\mathcal{G}^{F}$, respectively, where the edge structures are constructed based on sensor group knowledge and process flow knowledge. 
In this way, each attention module selectively reflects a specific style of process knowledge through its corresponding graph structure.

Using these aggregated feature vectors, we integrate all available information to predict the value of the $i$-th sensor at time $t$ as follows:
\begin{equation*}
    \hat{\textbf{x}}_{t,i} = \mathcal{F}\bigg(\textbf{a}_{t,i}^U \oplus \textbf{a}_{t,i}^{G} \oplus \textbf{a}_{t,i}^{F} \oplus e_1(i) \oplus e_2(i)\bigg),
\end{equation*}
where $\hat{\textbf{x}}_{t,i} \in \mathbb{R}$ is the $i$-th element of the predicted vector $\hat{\textbf{x}}_{t} \in\mathbb{R}^{N}$, and $\mathcal{F}$ is a multi-layer perceptron (MLP). 
During the training process, we minimize the mean squared error between $\mathbf{x}_t$ and $\hat{\mathbf{x}}_t$, utilizing it as the loss function:
\begin{equation*}
    \mathcal{L} = \frac{1}{T_\text{tr}-w}\sum\limits_{t=w+1}^{T_{\text{tr}}}||\hat{\mathbf{x}}_t -{\mathbf{x}}_t||_{2}^{2}.
\end{equation*}

\subsection{Anomaly Scoring}
\label{sec:anomaly scoring}
After training, we evaluate anomaly scores on the test dataset using the graph deviation scoring method \cite{deng2021graph}. 
This approach computes forecasting errors for each sensor, normalizes them, and defines the anomaly score as the maximum normalized error across sensors at each test timestamp.
Formally, at time $t$, the anomaly score $\mathcal{A}(t)$ is computed as:
\begin{equation*}
    \mathcal{A}(t) = \max_{i}\bigg(\frac{\text{Err}_i(t)-\tilde\mu_i}{\tilde\sigma_i}\bigg),
\end{equation*} 
where $i \in \{1,\ldots, N\}$ indexes the $N$ sensors, and $\text{Err}_{i}(t)=\lvert \hat{\mathbf{x}}_{t,i}-\mathbf{x}_{t,i} \rvert$ denotes the forecasting error for the $i$-th sensor at time $t$.
Here, $\mathbf{x}_{t,i} \in \mathbb{R}$ and $\hat{\mathbf{x}}_{t,i}\in\mathbb{R}$ represent the observed and predicted values, respectively. 
To ensure robustness against outliers, we normalize the error for each sensor using the median $\tilde{\mu}_{i}$ and interquartile range (IQR) $\tilde{\sigma}_{i}$ of its forecasting errors, following \cite{deng2021graph}. 
The median and interquartile range are computed from the training dataset, which contains only normal operating conditions.
Specifically, for each sensor $i$, $\tilde{\mu}_{i}$ is computed as the second quartile $Q_{2,i}$ of the training forecasting errors, and $\tilde{\sigma}_{i}$ is computed as $Q_{3,i}-Q_{1,i}$, where $Q_{1,i}$ and $Q_{3,i}$ denote the first and third quartiles, respectively.

A test timestamp $t$ is flagged as anomalous if $\mathcal{A}(t)$ exceeds a predefined threshold.

\section{Experiments}
\label{sec:experiments}

\subsection{Datasets}
We used the Secure Water Treatment (SWaT) \cite{mathur2016swat} and Water Distribution (WADI) \cite{ahmed2017wadi} datasets. The former was obtained from a multi-stage water treatment process, while the latter was obtained from a multi-stage water distribution process. Both datasets were collected by iTrust, a cybersecurity research center at the Singapore University of Technology and Design.

The SWAT dataset was collected from a six-stage water treatment process, as described in Section \ref{sec:introduction} and shown in Fig.~\ref{fig: figure 1}.  
It includes measurements from 51 sensors over 11 consecutive days, including 7 days of normal operation and 4 days during which cyber-attacks of varying time intervals were executed.

The WADI dataset was collected from a three-stage water distribution process (P1--P3) \cite{ahmed2017wadi}. 
In P1, the raw water is stored in two tanks, and its quality is controlled by a chemical dosing system. 
In P2, the water initially flows to elevated reservoir tanks and is subsequently distributed to consumer tanks based on a pre-set demand pattern. If the consumer tanks meet their demands, the water is reserved in another tank and sent back to P1, which is referred to as P3.
The WADI dataset comprises measurements of 127 sensors over 14 consecutive days of normal operations and 2 subsequent days of operations with several cyber attacks.

We provide the statistics for the SWaT and WADI datasets in Table I. We followed the data preprocessing procedures outlined in \cite{deng2021graph}. 
For both datasets, we used the data from normal operations for training and the data corresponding to cyber attack periods for testing. 
We used 10\% of the training data as a validation set for both datasets to tune hyperparameters and perform early stopping.

\begin{table}[hbtp!]
    \caption{Statistics for the SWaT and WADI datasets: number of sub-processes, number of sensors, number of training data points, number of testing data points, and the ratio of anomalies in the test data.}
    \label{table_1}
    \centering
    \begin{tabular}{c c c c c c}
    \hline
    \textbf{Datasets} &\textbf{\#Sub-Proc.} &\textbf{\#Sensors} & \textbf{\#Train} & \textbf{\#Test} &\textbf{\%Anomalies} \\ \hline 
    SWaT      &6  &51    &47520   &44991   &12  \\
    WADI      &3  &127   &118800  &17280   &6   \\ \hline
    \end{tabular}
\end{table}

\begin{table*}[htbp!]
\caption{Comparison of anomaly detection performance between the proposed method and various baselines in terms of \textbf{Pr}, \textbf{Re}, and \textbf{F1}. The values in bold indicate the highest scores. $\dagger$ denotes reproduced results using officially provided codes, while the remaining results of the baselines are sourced from \cite{jo2024edge}.}
\label{table_2}
\centering
\begin{tabular}{c c c c  | c c c }
\hline
&\multicolumn{3}{c|}{\textbf{SWaT}}  & \multicolumn{3}{c}{\textbf{WADI}} \\ \hline
\textbf{Method} &\textbf{Pr} & \textbf{Re} & \textbf{F1}  & \textbf{Pr} & \textbf{Re} & \textbf{F1}  \\ \hline
MAD-GAN   &0.9658 (0.0237)   &0.5750 (0.0135)  &0.7206 (0.0122)  &{0.8733} (0.0023)  &0.1487 (0.0004)  &0.2541 (0.0007)    \\
USAD    &\textbf{0.9941} (0.0003)    &0.5902 (0.0002)   &0.7406 (0.0002)   &\textbf{0.8743} (0.0028)   &0.1488 (0.0005)   &0.2543 (0.0008)   \\
TranAD  &0.5701 (0.4090)  &{0.7139} (0.1462)  &0.5003 (0.2332)  &0.8680 (0.0027)  &0.1478 (0.0005)  &0.2524 (0.0008)   \\ 
MTAD-GAT    &0.9515 (0.0253)   &0.6187 (0.0118)  &0.7495 (0.0030)  &0.6210 (0.2744)  &0.1857 (0.0333)  &0.2645 (0.0041) \\
GTA   &{0.9899} (0.0045)   &0.6411 (0.0295)  &0.7778 (0.0225)  &0.6195 (0.2094)  &0.3326 (0.0629)  &0.4083 (0.0502)  \\ 
GDN &0.8803 (0.1300) &0.6275 (0.0191) &0.7280 (0.0453)  &0.6625 (0.0925)  &0.3110 (0.0299)  &0.4206 (0.0362)  \\
ECNU-GNN  &0.9845 (0.0051) &0.6891 (0.0628)  &{0.8089} (0.0434)  &0.7652 (0.0684) &{0.4610} (0.0478) &{0.5730} (0.0455)  \\ 
FuSAGNet  &0.9776 (0.0104)$^\dagger$ &0.6422 (0.0133)$^\dagger$ &0.7750 (0.0100)$^\dagger$  &0.6521 (0.0940)$^\dagger$ &0.4550 (0.0415)$^\dagger$ &0.5301 (0.0313)$^\dagger$  \\
CAROTS  &0.9616 (0.0123)$^\dagger$ &0.6728 (0.0016)$^\dagger$  &{0.7916} (0.0036)$^\dagger$  &0.5446 (0.0816)$^\dagger$ &{0.4986} (0.0465)$^\dagger$ &{0.5150} (0.0216)$^\dagger$  \\ 
CGAD  &0.9662 (0.0300)$^\dagger$ &0.6166 (0.0160)$^\dagger$  &{0.7523} (0.0085)$^\dagger$ &0.2639 (0.0243)$^\dagger$ &{0.5356} (0.1101)$^\dagger$ &{0.3475} (0.0192)$^\dagger$  \\
TopoGDN  &0.9165 (0.0336)$^\dagger$ &0.6631 (0.0348)$^\dagger$  &{0.7683} (0.0207)$^\dagger$  &0.6283 (0.1038)$^\dagger$ &0.4102 (0.0380)$^\dagger$ &0.4899 (0.0166)$^\dagger$  \\ \hline
\textbf{Proposed} &0.9658 (0.0236) &\textbf{0.7188} (0.0387) &\textbf{0.8233} (0.0227)  &0.7576 (0.0173) &\textbf{0.5676} (0.0357) &\textbf{0.6485} (0.0260)  \\ \hline
\end{tabular}
\end{table*}

\begin{table*}[htbp!]
\caption{Comparison of anomaly detection performance between the proposed method and its variants in terms of \textbf{Pr}, \textbf{Re}, and \textbf{F1}. The values in bold indicate the highest scores.}
\label{tablee}
\centering
\begin{tabular}{c c c c  | c c c }
\hline
&\multicolumn{3}{c|}{\textbf{SWaT}}  & \multicolumn{3}{c}{\textbf{WADI}} \\ \hline
\textbf{Method} &\textbf{Pr} & \textbf{Re} & \textbf{F1}  & \textbf{Pr} & \textbf{Re} & \textbf{F1}  \\ \hline
\textbf{Proposed} &{0.9658} (0.0236) &\textbf{0.7188} (0.0387) &\textbf{0.8233} (0.0227) &\textbf{0.7576} (0.0173) &\textbf{0.5676} (0.0357) &\textbf{0.6485} (0.0260) \\
\textit{w/o} $\mathcal{G}^U$ &0.9681 (0.0188) &0.6932 (0.0366) &0.8072 (0.0246) &0.7475 (0.0341) &0.5005 (0.0458) &0.5979 (0.0296) \\
\textit{w/o} $\mathcal{G}^F$ &{0.9730} (0.0217) &{0.6807} (0.0456) &{0.8000} (0.0315) &{0.6962} (0.0599) &{0.5494} (0.0509) &{0.6111} (0.0360)\\
\textit{w/o} $\mathcal{G}^G$ &\textbf{0.9818} (0.0064) &{0.6651} (0.0518) &0.7918 (0.0356) &{0.7043} (0.0266) &0.5423 (0.0261) &{0.6123} (0.0185)\\
\textit{w/o} $\mathcal{G}^F, \mathcal{G}^G$ &{0.9807} (0.0102) &0.6445 (0.0326) &0.7774 (0.0253) &0.6979 (0.0472) &0.5148 (0.0214) &0.5918 (0.0237) \\
\textit{w/o} $\mathbf{e}_1, \mathbf{e}_2$ concat &0.9524 (0.0337) &0.6963 (0.0397) &0.8032 (0.0209) &0.7369 (0.0451) &0.4672 (0.0365) &0.5703 (0.0273) \\ \hline
\end{tabular}
\end{table*}

\subsection{Baselines}
We evaluate the proposed method against a diverse set of baselines, including recent state-of-the-art approaches, to ensure a fair and comprehensive comparison.
These baselines were carefully selected to cover various model architectures (e.g., LSTM, Transformer, GNN), anomaly scoring strategies (e.g., reconstruction-based, forecasting-based, hybrid), reproducibility under a common evaluation protocol, and the extent to which domain knowledge is incorporated. The selected baselines are briefly described below:

\begin{itemize}
    \item \textbf{MAD-GAN} \cite{li2019mad}: MAD-GAN is adversarially trained with a generator and a discriminator, both based on a long short-term memory (LSTM) network. The generator reconstructs input data, while the discriminator distinguishes between the input and reconstructed data. This method utilizes both reconstruction errors and discrimination scores to detect anomalies.
    \item \textbf{USAD} \cite{audibert2020usad}: USAD consists of two autoencoders that share a common encoder. It is trained within a two-phase adversarial framework and detects anomalies based on reconstruction errors.
    \item \textbf{TranAD} \cite{tuli2022tranad}: TranAD comprises two Transformer-based encoder-decoder networks. It is adversarially trained and utilizes reconstruction errors as anomaly scores, similar to USAD. However, unlike USAD, one of the encoder-decoder networks reconstructs the entire training time series to capture long-term temporal trends.  
    \item \textbf{MTAD-GAT} \cite{zhao2020multivariate}: MTAD-GAT employs two parallel GATs to capture inter-sensor and intra-sensor dependencies separately. The outputs of these GATs are then used as inputs for jointly optimized reconstruction and forecasting models. Both reconstruction errors and forecasting errors are utilized as anomaly scores.    
    \item \textbf{GTA} \cite{chen2021learning}: GTA is a Transformer model designed to capture inter-sensor dependencies through graph convolution. It learns a graph structure by using the Gumbel-softmax sampling technique and uses forecasting errors as anomaly scores.
    \item \textbf{GDN} \cite{deng2021graph}: GDN learns a graph to represent relationships between sensors by evaluating their similarities through sensor embeddings. It employs graph attention-based forecasting, using forecasting errors for anomaly detection. 
    \item \textbf{ECNU-GNN} \cite{jo2024edge}: ECNU-GNN employs an embedding for each sensor and learns a graph of sensors, similar to GDN. During graph convolution, for each sensor, the features of its neighboring sensors are transformed into new features based on the embedding of the central sensor. These transformed features are then aggregated to update the feature of the central sensor. To detect anomalies, forecasting errors are used as anomaly scores.
    \item \textbf{FuSAGNet} \cite{han2022learning}: FuSAGNet enhances GDN's robustness against noise by integrating a sparse autoencoder capable of extracting sparse latent representations of input data. These representations are used to reconstruct the input data and also serve as input for graph attention-based forecasting. It considers sensor group knowledge during sensor graph learning and uses both reconstruction errors and forecasting errors for anomaly detection.
    \item \textbf{CAROTS} \cite{pmlr-v267-kim25aa}: CAROTS integrates inter-sensor causality into contrastive learning for anomaly detection. It first extracts a causality matrix via a causal discovery model, and then employs two augmentors to generate causality-preserving and causality-disturbing samples, which serve as positives and negatives, respectively. An encoder is trained with a similarity-filtered one-class contrastive loss so that its latent space separates normal and abnormal samples based on causality. Anomaly scores are computed in the learned representation space.
    \item \textbf{CGAD} \cite{febrinanto2025entropy}: CGAD constructs a weighted causal graph among sensors using transfer entropy. It models the causal graph and temporal patterns with weighted graph convolutions and causal convolutions, and uses forecasting errors as anomaly scores.
    \item \textbf{TopoGDN} \cite{liu2024topogdn}: TopoGDN extends GDN by incorporating a multi-scale temporal convolution module to capture fine-grained temporal features and augmenting the graph attention mechanism with topological features derived from persistent homology. It learns a sensor graph via sensor embeddings and uses forecasting errors as anomaly scores.

\end{itemize}

\subsection{Evaluation Metrics}
We used precision (\textbf{Pr}), recall (\textbf{Re}), and F1 score (\textbf{F1}) to evaluate the anomaly detection performances of the proposed method and baselines, with \textbf{F1} serving as the primary evaluation metric. 
Precision measures the proportion of detected anomalies that are truly anomalous; thus, higher precision indicates fewer false alarms. 
Recall measures the proportion of true anomalies that are correctly detected; thus, higher recall indicates fewer missed anomalies. 
The F1 score is the harmonic mean of precision and recall and serves as a balanced summary when both false positives and false negatives are important. 
For all three metrics, higher values indicate better anomaly detection performance.

The metrics are defined as follows:
\begin{align*}
    \textbf{Pr} &= \frac{\text{TP}}{\text{TP}+\text{FP}}, \quad \textbf{Re} = \frac{\text{TP}}{\text{TP}+\text{FN}}, \\
    \textbf{F1} &= 2 \times \frac{\textbf{Pr} \times \textbf{Re}}{\textbf{Pr}+\textbf{Re}},
\end{align*}
where TP, FP, and FN denote the number of true positives, false positives, and false negatives, respectively. 
The threshold for anomaly detection was determined using a grid search method to maximize \textbf{F1}.

\subsection{Experimental Settings}
The proposed method was trained using the Adam optimizer \cite{kingma2014adam}. 
For the SWaT dataset, we used a learning rate of $5\times10^{-4}$ and a batch size of 32, while for the WADI dataset, we used a learning rate of $1\times10^{-3}$ and a batch size of 64. Training was conducted for up to 200 epochs, with early stopping applied after 5 epochs of no improvement. 
The key hyperparameters $(k, \alpha, d)$ were selected through a validation-based joint grid search using validation F1 score as the model-selection criterion.
Specifically, we considered $\alpha \in \{0.1,0.3, 0.6,0.9\}$, $k \in \{5,10,15,20,25\}$, and $d \in \{32,64,128,256\}$, and evaluated all combinations jointly. 
The selected settings were $k=10$, $\alpha=0.6$, $d=128$ for SWaT and $k=10$, $\alpha=0.3$, and $d=128$ for WADI. For both datasets, the sliding window size $w$ was set to 15.

For the network architecture, including the number of layers and the TCN configuration, we adopted commonly used design choices from the TCN literature \cite{van2016wavenet,bai2018empirical}, and evaluated small variations around them. 
Deeper variants were not considered because they were unnecessarily over-parameterized for the input dimensionality of the SWaT and WADI benchmarks. 
The final architecture was selected based on the validation F1 score.
We employed a TCN with three dilated causal convolutional layers as the temporal encoder, where the dilation factors were set to 1, 2, and 4, respectively, and the filter size was set to 3. We used an MLP consisting of 1 layer and 2 layers with a hidden dimension of 128 for SWaT and WADI, respectively.
Sensor group knowledge and process flow knowledge were obtained from the process flow diagrams of the underlying processes for the SWaT and WADI datasets, accessible at the iTrust website (\url{https://itrust.sutd.edu.sg}).

\subsection{Experimental Results}

\subsubsection{Comparison of Performance with Baselines}
To demonstrate the effectiveness of the proposed method, we compared its anomaly detection performance against baselines. Table II summarizes the results on the SWaT and WADI datasets. We present the average \textbf{Pr}, \textbf{Re}, and \textbf{F1} over 10 repeated experiments, with values in parentheses indicating standard deviations.

The proposed method achieved the highest average \textbf{F1} scores on both SWaT (0.8233) and WADI (0.6485), consistently outperforming all baselines. 
This superior performance can be attributed to the explicit incorporation of two types of process knowledge---sensor group knowledge and process flow knowledge---into the graph structure learning stage.
By embedding these domain insights, the model can more effectively capture complex dependencies across sensors and sub-processes, which purely data-driven approaches often fail to recover, leading to improved anomaly detection performance.

We acknowledge that our method leverages additional knowledge not utilized by existing baselines, which may limit the fairness of a direct comparison. 
However, our goal is not to claim superiority under identical assumptions, but rather to demonstrate the value of incorporating process knowledge---often readily available in real-world settings---for enhancing anomaly detection.
Importantly, most baseline models are not structurally designed to incorporate such knowledge, even in simplified forms, making direct adaptation in our experiments non-trivial.
As a result, we evaluate the baselines in their original form without process knowledge and instead focus on demonstrating the benefits of our knowledge-guided approach.

\subsubsection{Ablation Study}
To verify the contribution of each component in the proposed framework, we conducted an ablation study including not only the knowledge-informed graphs but also the uninformed graph and the concatenation of the learnable embeddings. 
Specifically, we compared the proposed model with five variants: removing the uninformed graph (w/o $\mathcal{G}^{U}$), removing the process-flow-informed graph (w/o $\mathcal{G}^{F}$), removing the sensor-group-informed graph (w/o $\mathcal{G}^{G}$), removing both knowledge-informed graphs (w/o $\mathcal{G}^{F}, \mathcal{G}^{G}$), and removing the concatenation of $e_{1}$ and $e_{2}$ (w/o $e_{1}, e_{2}$ concat).

As shown in Table III, the performance consistently declined when each knowledge component was removed. 
Removing either $\mathcal{G}^{F}$ or $\mathcal{G}^{G}$ led to noticeable drops in average \textbf{F1}, and excluding both further degraded performance on both datasets.
These results confirm that sensor group knowledge and process flow knowledge each contribute independently to modeling inter-sensor dependencies, and that their combination yields synergistic improvements in anomaly detection performance.
The variants without $\mathcal{G}^{U}$ and without the concatenation of $e_{1}$ and $e_{2}$ further show that the uninformed graph provides complementary data-driven relational information beyond the knowledge-informed graphs, while the learnable embeddings contribute useful information to the prediction module in addition to their role in graph construction.

\subsubsection{Qualitative Analysis}

To gain a deeper understanding of the model's behavior beyond the experimental results, we present a qualitative analysis of two representative attack scenarios from the SWaT dataset.

\begin{figure}[!t]
    \centering
    \subfloat[]{\includegraphics[width=\columnwidth]{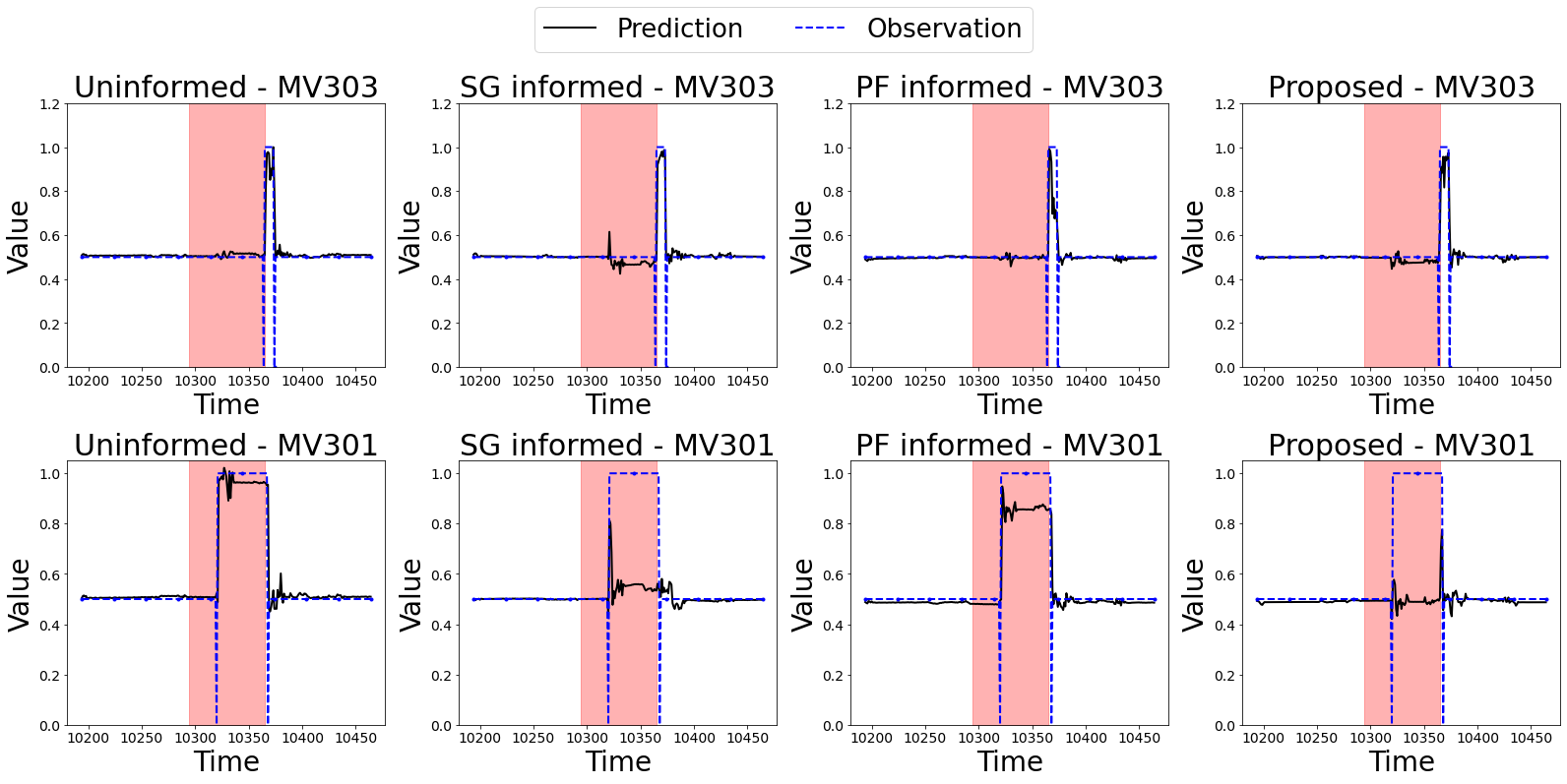}\label{fig: case 1a}}\\
    \subfloat[]{\includegraphics[width=\columnwidth]{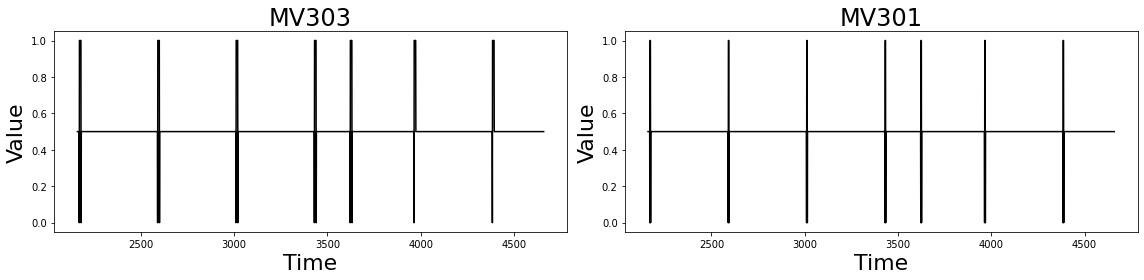}\label{fig: case 1b}}
    \caption{(a) Predicted and observed values during the attack on MV303 for the uninformed, SG-informed, PF-informed, and proposed models. The attack period is shaded in red. (b) Multivariate time series of sensors MV303 and MV301 in P3 under normal conditions.}
    \label{fig: case1}
\end{figure}

\textbf{Case 1: Intra-Sub-process Dependency (Attack on MV303)}
Fig.~\ref{fig: case1} illustrates an attack scenario in sub-process P3, where MV303 (a valve controller) is prevented from opening during the attack period. The attack creates a mismatch between MV303 and other sensors within the same sub-process, particularly MV301, another valve controller in P3 that regulates the backwashing process for the ultra-filtration unit.
Ideally, a robust MTAD model should detect such intra-sub-process inconsistencies by leveraging structural knowledge that sensors within the same sub-process tend to operate in a coordinated manner. In this case, when MV303 behaves abnormally while MV301 continues to follow its normal pattern, the model should flag the resulting deviation through elevated prediction errors.

In Fig.~\ref{fig: case 1a}, we compare the prediction results of four model variants during the attack period (shaded in red):
\begin{itemize}
\item \textbf{Uninformed Model}: Uses only data-driven graph learning without any process knowledge
\item \textbf{SG-Informed Model}: Incorporates sensor group knowledge to capture intra-sub-process relationships
\item \textbf{PF-Informed Model}: Incorporates process flow knowledge to capture inter-sub-process relationships
\item \textbf{Proposed Model}: Combines both types of process knowledge
\end{itemize}

The key observation is that both the SG-informed model and the proposed model exhibit significantly larger prediction errors for MV301 during the attack period. This suggests that these models have captured the dependency between MV303 (the attacked sensor) and MV301 (a related sensor within the same sub-process).
This behavior is plausible because the SG-informed model explicitly incorporates intra-sub-process relationships through sensor group knowledge, allowing the anomaly at MV303 to influence the prediction of MV301. In contrast, the uninformed model and the PF-informed model do not explicitly encode intra-sub-process relationships and thus fail to reflect the effect of MV303's anomaly on MV301, resulting in smaller prediction errors and missed anomaly detection. 
The strong correlation between MV303 and MV301 under normal operating conditions, as shown in Fig.~\ref{fig: case 1b}, suggests that these sensors exhibit coordinated behaviors within the same sub-process. This correlation likely reflects an underlying functional dependency, as both valves participate in the backwashing operation of the ultra-filtration unit.

\begin{figure}[!t]
    \centering
    \subfloat[]{\includegraphics[width=\columnwidth]{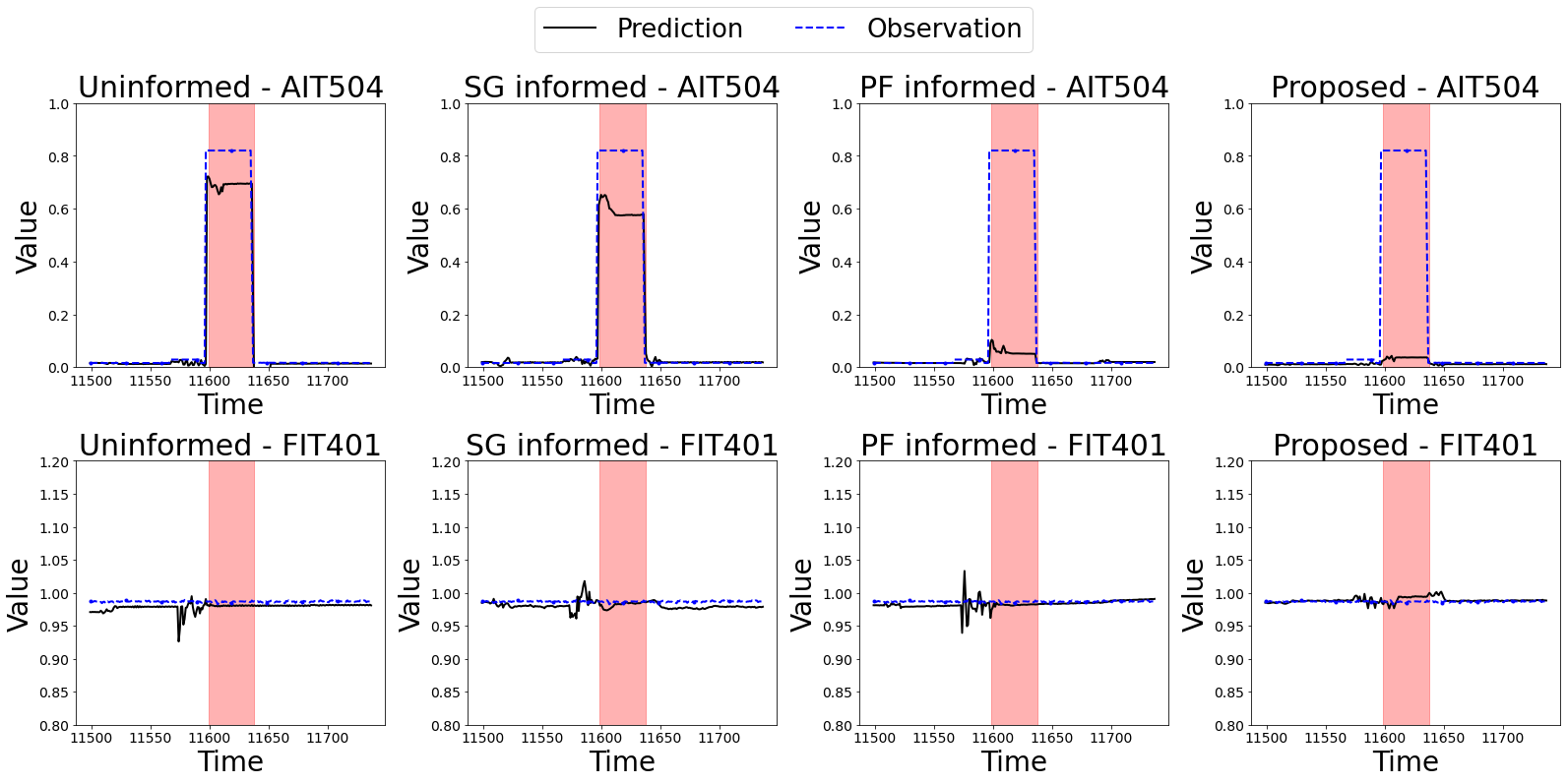}\label{fig: case 2a}}\\
    \subfloat[]{\includegraphics[width=\columnwidth]{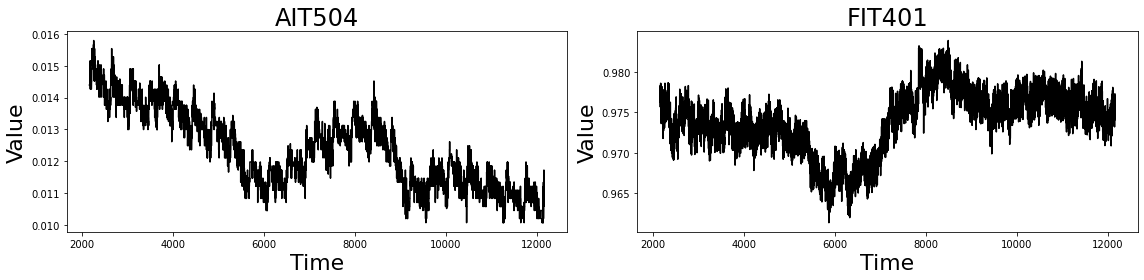}\label{fig: case 2b}}
    \caption{(a) Predicted and observed values during the attack on AIT504 for the uninformed, SG-informed, PF-informed, and proposed models. The attack period is shaded in red. (b) Multivariate time series of sensors AIT504 in P5 and FIT401 in P4 under normal conditions.}
    \label{fig: case2}
\end{figure}

\begin{figure*}[!t]
    \centering
    \subfloat[]{\includegraphics[width=0.32\linewidth]{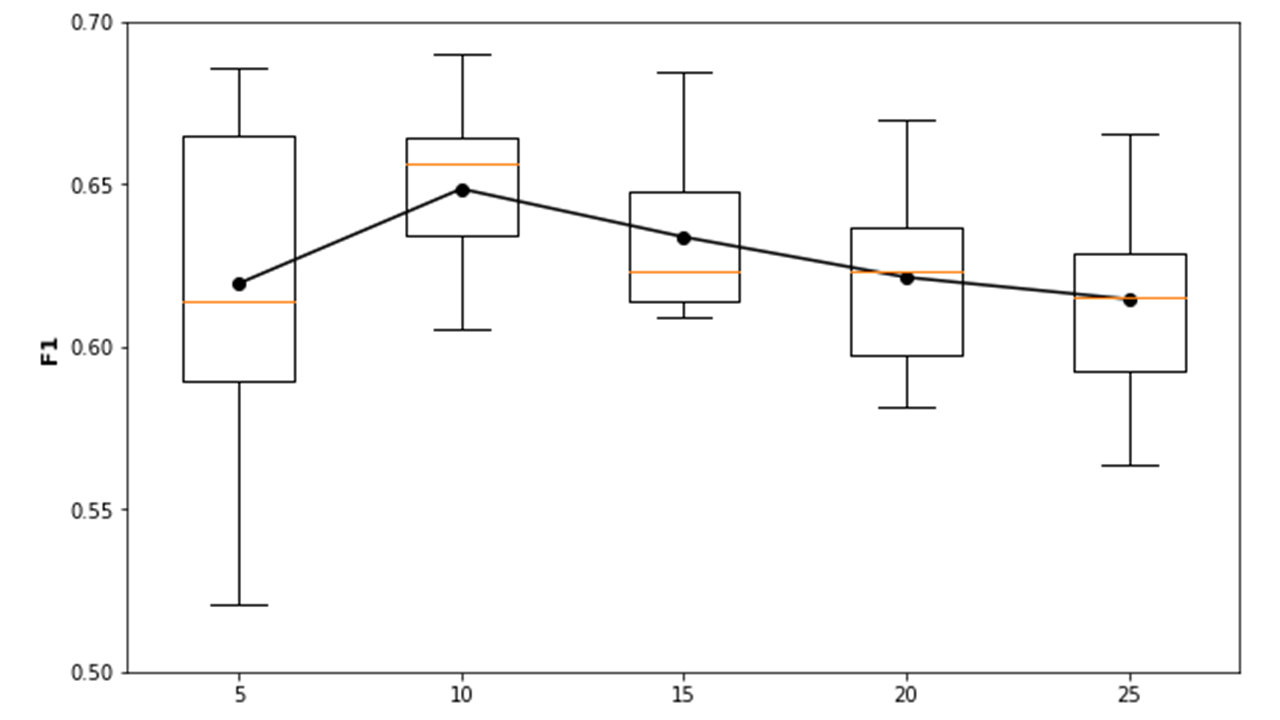}\label{fig: figure 3a}}
    \subfloat[]{\includegraphics[width=0.32\linewidth]{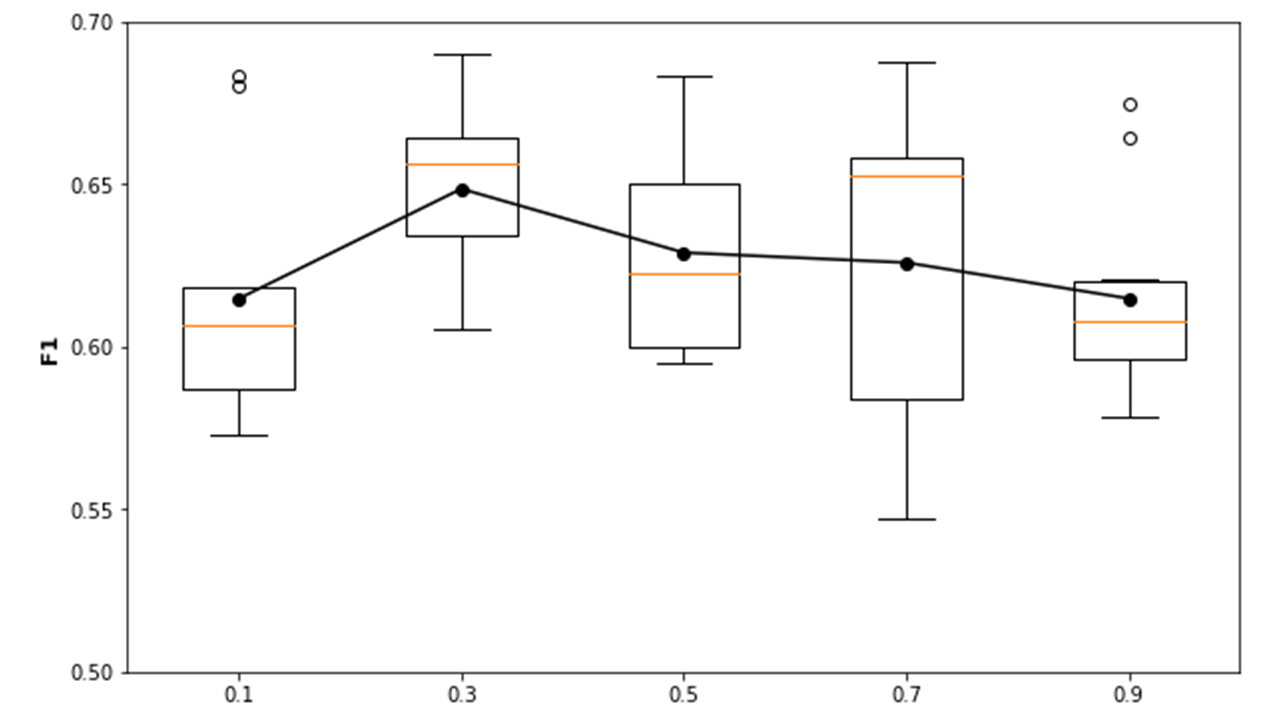}\label{fig: figure 3b}}
    \subfloat[]{\includegraphics[width=0.32\linewidth]{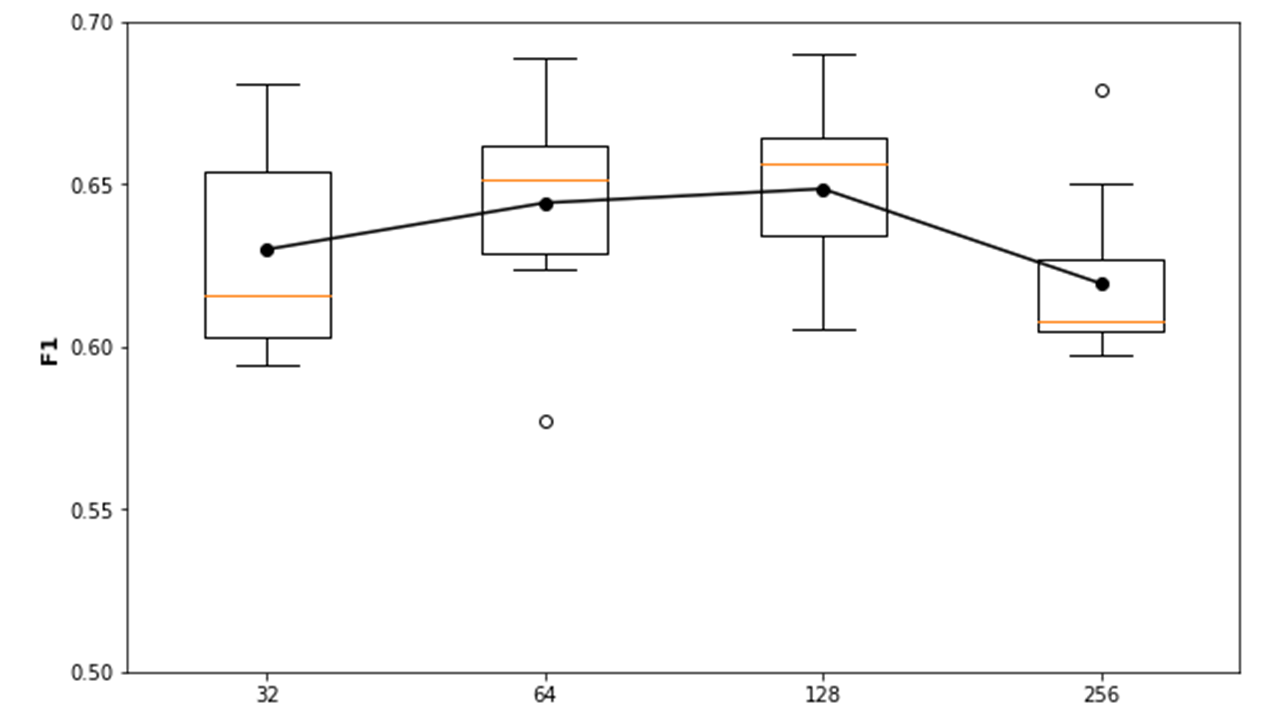}\label{fig: figure 3c}}
    \caption{Results of the hyperparameter sensitivity analysis: (a) $k$; (b) $\alpha$; (c) $d$. Black solid circles in the boxes indicate the mean values.}
    \label{fig: figure 3}
\end{figure*}

\textbf{Case 2: Inter-Sub-process Dependency (Attack on AIT504)}
Fig.~\ref{fig: case2} presents another attack scenario targeting AIT504 that measures water quality parameters in sub-process P5. The attack artificially manipulates the sensor value above its actual value, causing inconsistencies with related measurements in adjacent sub-processes. 
Ideally, a robust MTAD model should detect such inconsistencies by leveraging known dependencies---for example, recognizing that AIT504 is typically aligned with FIT401. 
If FIT401 remains stable during the attack, the model should continue to predict normal values for AIT504, resulting in a large prediction error when its observation is manipulated.

Fig.~\ref{fig: case 2a} demonstrates how different models respond to this attack.
The {PF-informed model} and {proposed model} demonstrate markedly higher forecasting errors for AIT504 during the attack period. As shown in the plots, these models predict values for AIT504 that are much lower than the manipulated readings and are closer to what the true (unmanipulated) values are expected to be. This leads to significant discrepancies between prediction and observation, resulting in large discrepancies that enable reliable anomaly detection.
This outcome highlights the effectiveness of explicitly incorporating inter-sub-process dependencies through process flow knowledge. 
In particular, FIT401 is a sensor located in sub-process P4, which directly influences sub-process P5, where AIT504 is located. Process flow knowledge explicitly encodes such inter-sub-process relationships, allowing the PF-informed and the proposed models to leverage this dependency during prediction. 
Since FIT401 remains unaffected by the attack and continues to exhibit normal behavior during the attack period, the PF-informed and the proposed models utilize its stable pattern to infer that AIT504 should also remain within a normal range. As a result, despite the manipulated observations of AIT504, the models predict its normal (unmanipulated) values. This leads to large prediction errors during the attack period, thereby enabling effective anomaly detection.
In contrast, the {uninformed model} and the {SG-informed model} lack this structural guidance since they do not explicitly encode such inter-sub-process relationships. Consequently, they tend to overfit to the anomalous (manipulated) observations of AIT504 more closely, yielding lower prediction errors and failing to detect the anomaly effectively.

Fig.~\ref{fig: case 2b} further supports this interpretation by showing that AIT504 and FIT401 exhibit similar trends under normal conditions---consistent with their connection in the process flow. 
The PF-informed and the proposed models effectively leverage this knowledge to identify anomalies that break such cross-stage consistency.

From this qualitative analysis, we derive the following observations based on two representative attack scenarios. 
First, different types of process knowledge play complementary roles in anomaly detection. In Case 1, incorporating sensor group knowledge helps the model capture intra-sub-process relationships, enhancing sensitivity to anomalies affecting related sensors. In Case 2, leveraging process flow knowledge allows the model to account for inter-sub-process dependencies, which is particularly beneficial when anomalies cause inconsistencies across adjacent process stages.

Second, these observations suggest that knowledge-informed structures can guide the model to focus on physically meaningful relationships---dependencies that may not be easily inferred from data alone---thereby improving the discovery of underlying dependency structures.

Overall, these findings qualitatively support the design of our proposed model, which integrates both sensor group and process flow knowledge. The results indicate that this approach leads to more robust anomaly detection performance in complex industrial systems.

\subsubsection{Hyperparameter Sensitivity Analysis}
We conducted a hyperparameter sensitivity analysis using the WADI dataset to investigate the impact of key hyperparameters, including $k$, $\alpha$, and $d$, using a one-factor-at-a-time protocol around the selected configuration. Specifically, when varying one hyperparameter, the other two were fixed to the values selected by the validation-based joint grid search.
When varying $k$ over $\{5,10,15,20,25\}$, we fixed $\alpha=0.3$ and $d=128$; when varying 
$\alpha$ over $\{0.1,0.3,0.5,0.7, 0.9\}$ we fixed $k=10$ and $d=128$; and when varying $d$ over $\{32,64,128,256\}$, we fixed $k=10$ and $\alpha=0.3$.
This analysis is intended to assess local robustness around the selected configuration rather than fully characterize higher-order interactions among hyperparameters.
Especially, as shown in Fig.~\ref{fig: figure 3}(a) and Fig.~\ref{fig: figure 3}(b), a decrease in performance was observed when $k$ and $\alpha$ were too small, possibly because the learned graphs were too sparse to capture complex inter-sensor dependencies. Conversely, performance also declined when these parameters were excessively large, likely due to redundant edges within the graphs, impeding effective capture of inter-sensor dependencies.

\begin{table}[t!]
\caption{Average forecasting errors on normal and anomalous timestamps in the test set, measured by MAE and RMSE. 
Values are averaged over 10 repeated experiments, with standard deviations shown in parentheses.}
\label{table_forecast}
\centering
\resizebox{\columnwidth}{!}{
\begin{tabular}{c c c | c c}
\hline
& \multicolumn{2}{c|}{\textbf{SWaT}} & \multicolumn{2}{c}{\textbf{WADI}} \\ \hline
& \textbf{MAE} & \textbf{RMSE} & \textbf{MAE} & \textbf{RMSE} \\ \hline
Normal & 0.0527 (0.0098) & 0.2717 (0.0120) & 0.0124 (0.0038) & 0.1362 (0.0766) \\
Anomalous & 0.2450 (0.0449) & 0.4617 (0.0464) & 0.3522 (0.3270) & 3.7431 (1.5384) \\ \hline
\end{tabular}
}
\end{table}

\subsubsection{Forecasting Error Analysis}
To directly assess the forecasting accuracy of the proposed model under normal operating conditions, which underpins the prediction-error-based anomaly detection mechanism, we computed the mean absolute error (MAE) and root mean squared error (RMSE) separately on normal and anomalous timestamps in the test set. 
As shown in Table~\ref{table_forecast}, the forecasting errors on normal timestamps are substantially lower than those on anomalous timestamps for both datasets.
These results indicate that the proposed method maintains low forecasting errors under normal operating conditions, while anomaly timestamps yield markedly larger prediction errors.
This clear separation demonstrates that forecasting error serves as an effective discriminative signal for distinguishing anomalous from normal observations, thereby validating the effectiveness of the forecasting error-based anomaly scoring mechanism.

\section{Conclusion}
\label{sec:conclusion}
We propose a process knowledge-assisted multi-graph dependency learning framework for GNN-based MTAD tailored for multi-stage industrial processes.
The core contribution lies in enhancing graph structure learning by explicitly incorporating two forms of domain knowledge: sensor group knowledge and process flow knowledge. 
This structured integration helps the model to more effectively capture inter-sensor dependencies, improving over conventional data-driven approaches that often overlook the underlying process dynamics. 
By coupling this knowledge-guided graph structure construction with temporal modeling via TCN, the proposed method delivers more robust forecasting and improved anomaly detection performance on real-world industrial datasets.

A notable limitation of the current method is that it does not explicitly account for the inherent stochasticity in industrial time series data, which may lead to overconfident predictions and reduced anomaly detection robustness.
As a future direction, we plan to extend our process knowledge-guided graph learning framework into a probabilistic forecasting-based MTAD model. 
By modeling forecast uncertainty alongside structured graph-based dependencies, this extension seeks to enhance anomaly detection reliability while retaining the core strengths of our current approach.

\section*{Acknowledgment}
This research was supported by the National Research Foundation of Korea (NRF) grant funded by the Korea government (MSIT) (2023R1A2C2005453, RS-2023-00218913).

\end{document}